# Decoding Funded Research: Comparative Analysis of Topic Models and Uncovering the Effect of Gender and Geographic Location

Shirin Tavakoli Kafiabad[1], Andrea Schiffauerova[1], and Ashkan Ebadi[1,2,*]

[1] Concordia Institute for Information Systems Engineering, Concordia University, Montreal, QC H3G 2W1 Canada
[2] National Research Council Canada, Toronto, ON M5T 3J1, Canada

[*] Email: ashkan.ebadi@nrc-cnrc.gc.ca

**Abstract** Optimizing national scientific investment requires a clear understanding of evolving research trends and the demographic and geographical forces shaping them, particularly in light of commitments to equity, diversity, and inclusion. This study addresses this need by analyzing 18 years (2005–2022) of research proposals funded by the Natural Sciences and Engineering Research Council of Canada (NSERC). We conducted a comprehensive comparative evaluation of three topic modelling approaches: Latent Dirichlet Allocation (LDA), Structural Topic Modelling (STM), and BERTopic. We also introduced a novel algorithm, named COFFEE, designed to enable robust covariate effect estimation for BERTopic. This advancement addresses a significant gap, as BERTopic lacks a native function for covariate analysis, unlike the probabilistic STM. Our findings highlight that while all models effectively delineate core scientific domains, BERTopic outperformed by consistently identifying more granular, coherent, and emergent themes, such as the rapid expansion of artificial intelligence. Additionally, the covariate analysis, powered by COFFEE, confirmed distinct provincial research specializations and revealed consistent gender-based thematic patterns across various scientific disciplines. These insights offer a robust empirical foundation for funding organizations to formulate more equitable and impactful funding strategies, thereby enhancing the effectiveness of the scientific ecosystem.

## 1. Introduction

In today's rapidly evolving global landscape, understanding the forces that drive scientific inquiry is more crucial than ever. As global investment in science and technology continues to increase (Bloch and Sørensen 2014; Ebadi et al. 2020), there is a growing imperative for precise, data-driven insights into research trends and funding dynamics. These insights are essential to ensure that substantial investments translate into meaningful social and economic returns and contribute to equitable resource allocation (Stephan 2012).

While identifying which research initiatives are receiving funding is crucial, gaining a comprehensive understanding of who conducts this research and where it takes place provides a more holistic view (Bornmann, Mutz, and Daniel 2007; Urquhart-Cronish, Otto, and Ogden 2019; Witteman et al. 2019). The existing literature highlights persistent gender and geographic disparities in scientific funding and participation (Breschi and Lissoni 2010; Grillitsch, Fitjar, and Stambøl 2019; Rodriguez-Pose et al. 2019). However, there is a critical gap in systematically quantifying how these researcher characteristics influence the prevalence of specific research topics within large-scale funding datasets, and how these relationships might change over time, potentially reflecting systemic inequities or shifts in research priorities.

To effectively navigate and extract meaningful insights from extensive textual funding datasets, in this work, we employ advanced computational techniques in Natural Language Processing (NLP) and Artificial Intelligence (AI). Our focus is on topic modelling (TM), a powerful tool that enables uncovering the latent themes embedded within the corpus and systematically tracking their evolution across demographic and temporal dimensions (Blei and Lafferty 2006; Rosen-Zvi et al. 2004). Methodologically, first, we undertake a rigorous comparative analysis of three prominent topic models: 1) Latent Dirichlet Allocation (LDA) (Blei 2003), which utilizes probabilistic inference to identify patterns in the data, 2) Structural Topic Modelling (STM) (Roberts et al. 2013), which builds on LDA by incorporating document-level covariates, allowing for a more nuanced understanding of thematic variations influenced by external factors, and 3) an advanced topic modelling approach based on Bidirectional Encoder Representations from Transformers (BERT), which leverages transformer-based contextual embeddings to capture semantic nuances (Devlin et al. 2019; Grootendorst 2020). This multi-model comparison is crucial as it enables us to assess how different underlying modelling assumptions, ranging from probabilistic inference in LDA and STM to semantic space analysis of the BERT-based approach (Bianchi, Terragni, and Hovy 2021), affect the coherence of identified topics and the interpretability of subsequent downstream analysis.

Next, we conduct a comparative estimation of the quantitative influence of researcher characteristics, particularly gender and geographical locations, on the prevalence of different research topics in both the STM and BERTopic. To achieve this, we treat topic prevalence as a continuous outcome and apply robust regression-based effect estimation techniques. For the STM model, we use the established *estimateEffect* function of the *stm* package in R (Roberts, Stewart, and Tingley 2019), which is specifically designed to incorporate topic-model uncertainty through variational posterior sampling. Recognizing that the BERTopic approach, by its nature, is not a generative probabilistic model and thus lacks native functions for direct effect estimation with integrated uncertainty, in this work, we introduce a novel algorithm, the Covariate Effect Estimation for the BERTopic model (*COFFEE*), implemented in *Python*. This innovative approach emulates the STM framework by generating multiple bootstrapped topic distributions and subsequently applying Ordinary Least Squares (OLS) regression to robustly estimate the effects of demographic and geographic covariates. Although this strategy does not replicate the full probabilistic machinery of STM, it critically enables meaningful methodological comparisons by capturing uncertainty through nonparametric resampling (Efron and Tibshirani 1994; Mooney 1996) and aligning the structure of the statistical analysis across distinct modelling paradigms (Egger and Yu 2023).

This study is driven by two central research questions: First, what are the latent research themes within the Canadian funding landscape from 2005 to 2022, and how have these themes evolved over time? Second, how do researcher characteristics, specifically gender and geographical location, influence the prevalence of these identified research topics? By addressing these questions, our work makes several significant contributions to the understanding of the Canadian STEM research ecosystem. We provide a comprehensive analysis of eighteen years of publicly available funded research proposals, employing three advanced topic modelling techniques. The development of the novel COFFEE algorithm for BERTopic, combined with the use of built-in effect estimate functionalities in STM, allows for a robust quantification of the impact of researcher demographics on topic prevalence. This methodological advancement is particularly timely and relevant, aligning with national initiatives like the Tri-Agency Equity, Diversity, and Inclusion (EDI) action plan (NSERC 2021), which emphasizes the importance of fostering an inclusive and representative research environment across Canada. Our findings not only enhance the empirical understanding of funding dynamics but also contribute to providing actionable insights for policy-makers and stakeholders to develop more equitable and impactful funding strategies that can better support a diverse research community.

The remainder of this paper is structured as follows: Section 2 presents a comprehensive review of relevant literature. Section 3 details the data sources and the preprocessing step. Our methodology, encompassing topic modelling approaches, comparative evaluation metrics, and effect estimation techniques, is elaborated in Section 4. Section 5 presents our key findings, interprets the identified topics, and discusses the estimated effects of geographical and gender covariates on the research theme. Finally, Section 6 offers our concluding remarks and discussions, and Section 7 discusses the limitations of this study and outlines directions for future research.

## 2. Literature Review

### 2.1. Research Funding and Its Impact

Investing in research and development is a critical driver of innovation and economic sustainability (Ebadi et al. 2020; Stephan 2012). Understanding the downstream impact of grant funding on scientific productivity is crucial for optimizing resource allocation and maximizing societal benefits (Ebadi and Schiffauerova 2016a; Jacob and Lefgren 2011). This necessitates a comprehensive understanding of evolving research trends for effective and equitable allocation of resources. Recent analyses of global R&D expenditure highlight the increasing prioritization of science and technology as tools for addressing societal challenges, though disparities in funding distribution persist across regions and demographics (Bloch and Sørensen 2014; Ebadi et al. 2020). In Canada, initiatives such as the Tri-Agency Equity, Diversity, and Inclusion (EDI) Action Plan (NSERC 2021) reflect a growing recognition of systemic inequities in research ecosystems, underscoring the need for data-driven assessments of funding patterns.

### 2.2. Gender Disparities in Scientific Research

The role of demographic factors, such as gender, in shaping research priorities remains relatively less explored. Studies have consistently demonstrated gender disparities within scientific research (Abramo, D'Angelo, and Di Costa 2019; van den Besselaar and Mom 2021). For instance, Larivière et al. (Larivière et al. 2013) highlighted global gender disparities, indicating that women are often underrepresented in scientific authorship. Similarly, van Arensbergen et al. (van Arensbergen, van der Weijden, and van den Besselaar 2014) found that gender differences in scientific productivity and funding success persist across various disciplines, emphasizing the continued need for more inclusive approaches to funding allocation. In Canada, Else et al. (Else et al. 2020) found that factors predicting research grant success vary, suggesting the potential influence of gender. Moreover, persistent gender gaps in Canadian STEM funding, particularly in engineering and physical sciences, have been documented, showing, for instance, higher rejection rates for early-career women scientists and slightly less funding for successful female applicants (Goldin and Urquhart-Cronish 2019). Hajibabaei et al. (Hajibabaei, Schiffauerova, and Ebadi 2022, 2023) found that in AI research networks. At the same time, scientific performance is key for both genders to achieve central roles. Women face subtle disadvantages in possessing core positions, which further highlights ongoing gender disparities (Hajibabaei et al. 2023). These disparities are potentially exacerbated by biases in evaluation processes, as evidenced by studies suggesting that grant applications led by women may be evaluated less favorably than those led by men, even when scientific quality is comparable (Bornmann et al. 2007; Wennerås and Wold 1997; Witteman et al. 2019).

## 2.3. Geographical Disparities in Research Funding

Geographical disparities in research funding allocation remain a persistent challenge, mirroring demographic inequities. Breschi and Lissoni (Breschi and Lissoni 2010) illustrate how regional proximity and inventor mobility drive knowledge spillovers, emphasizing the role of geography in shaping innovation. The concentration of funding in metropolitan hubs can marginalize researchers in peripheral regions, limiting access to resources and collaboration opportunities (Grillitsch et al. 2019; Rodriguez-Pose et al. 2019). This spatial inequity aligns with the "Matthew Effect" in science described by Merton (Merton 1968), where established centers attract disproportionate resources, exacerbating regional imbalance. Broader studies on regional innovation systems (Grillitsch et al. 2019) and the socio-economic impacts of regional inequality (Rodriguez-Pose et al. 2019) highlight the importance of considering geographical factors in science policy.

## 2.4. Topic Modelling

Topic modelling has emerged as a crucial tool for analyzing large text datasets to identify patterns and recurring themes. The foundation for modern topic modelling was established by Blei et al. (Blei 2003), who introduced Latent Dirichlet Allocation (LDA). LDA, a generative probabilistic model, has since become a fundamental method for text analysis. Building upon LDA, Structural Topic Models (STM) (Roberts et al. 2013) enable researchers to incorporate metadata (e.g., authorship demographics, temporal trends) directly into topic discovery, a feature critical for analyzing the interplay between research content and contextual factors (Roberts et al. 2019). The emergence of transformer-based architectures, such as BERT (Devlin et al. 2019), has recently revolutionized natural language processing by enabling models to capture intricate contextual relationships within text. Building upon this, transformer-based topic models, e.g., BERTopic (Grootendorst 2020), leverage advanced contextual embeddings to capture semantic relationships in text, showing strong performance for analyzing complex documents (Bianchi et al. 2021). Comparative studies of LDA, STM, and BERTopic emphasize trade-offs between probabilistic interpretability and contextual nuance (Egger and Yu 2023), a methodological tension central to our study. Evaluating topic models involves assessing topic coherence, which quantifies the interpretability of topics (Mimno et al. 2011; Röder, Both, and Hinneburg 2015), particularly relevant for models such as BERTopic. The underlying dimensionality reduction techniques used in some models, e.g., BERTopic, often rely on algorithms such as Uniform Manifold Approximation and Projection (UMAP) (Healy and McInnes 2024), while clustering is performed with methods like HDBSCAN (McInnes, Healy, and Astels 2017), highlighting the mixed nature of these approaches.

## 2.5. Methodological Approaches to Effect Estimation

The application of regression techniques to topic model outputs for effect estimation builds upon a growing body of work in computational social science and text analysis. For instance, studies have utilized models that link author characteristics to topics discussed in scientific literature (Rosen-Zvi et al. 2004) and used regression to understand how temporal trends affect thematic shifts (Blei and Lafferty 2006). Our approach extends this line of inquiry by specifically focusing on gender and geographic covariates within the context of publicly funded research proposals in Canada. By employing and adapting these effect estimation techniques for both STM and BERTopic models, we aim to provide a more comprehensive understanding of the factors shaping the landscape of Canadian research funding. The application of topic modelling to understand research trends and inform science policy has been explored in studies tracking the dynamics of research topics and comparing different topic modelling methods for analyzing scientific literature (Boyack and Klavans 2011; Ebadi et al. 2020). By integrating these methodological and empirical foundations, this study bridges gaps in understanding how gender and geographical factors intersect with funded research projects.

## 3. Data

In this study, we utilized a comprehensive dataset of scientific projects and research proposals funded by the Natural Sciences and Engineering Research Council of Canada (NSERC), spanning the years 2005 to 2022. The final dataset contained 78,863 proposal summaries. This publicly available dataset includes key features such as: Application ID, Application Title, First name and last name of the Researcher, Organization, Province, Country, Competition Type, Award Amount, and Application summary.

To determine the gender of the researchers, we employed a multifaceted approach using the GPT-4 model (OpenAI Team et al. 2024). For each researcher, GPT-4 was prompted with the researcher's first name, last name, and country of origin to classify the gender as "male", "female", or "ambiguous". The prompt was carefully engineered to encourage GPT-4 to utilize its extensive training data, which includes implicit linguistic patterns (e.g., gendered name suffixes in various languages) and the regional prevalence of names. In instances where GPT-4 classified a name as "ambiguous", we subsequently adopted the most frequently used gender associated with that name historically. For example, if "Taylor" was historically more commonly associated with males, it was classified as male in our dataset.

To ensure the accuracy of our gender classifications, we conducted a thorough validation process. A stratified random sample of 100 researchers was manually checked against established records and common knowledge. This verification process revealed an overall accuracy rate of 93%, with a breakdown showing 91% accuracy for female classifications and 95% accuracy for male classifications. For names initially classified as "ambiguous" by GPT-4 and subsequently resolved using historical frequency data, our accuracy was confirmed at 88%. The resulting gender distribution is shown in Figure 1.

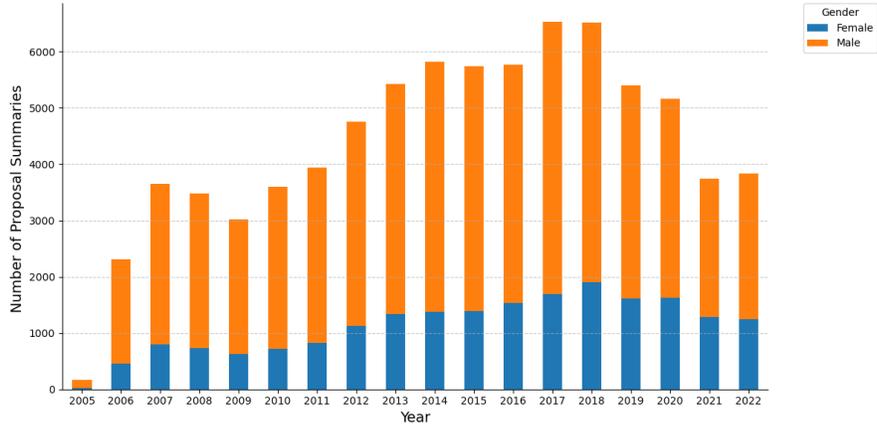

**Figure 1.** Distribution of gender in the proposal summaries over time.

## 4. Methodology

Our methodology consists of four key components, as shown in Figure 2: (1) Data collection and pre-processing, (2) Topic modelling, (3) Comparative analysis and evaluation, and (4) Effect estimation of geographical locations and gender. The following subsections describe each of these key components in more detail.

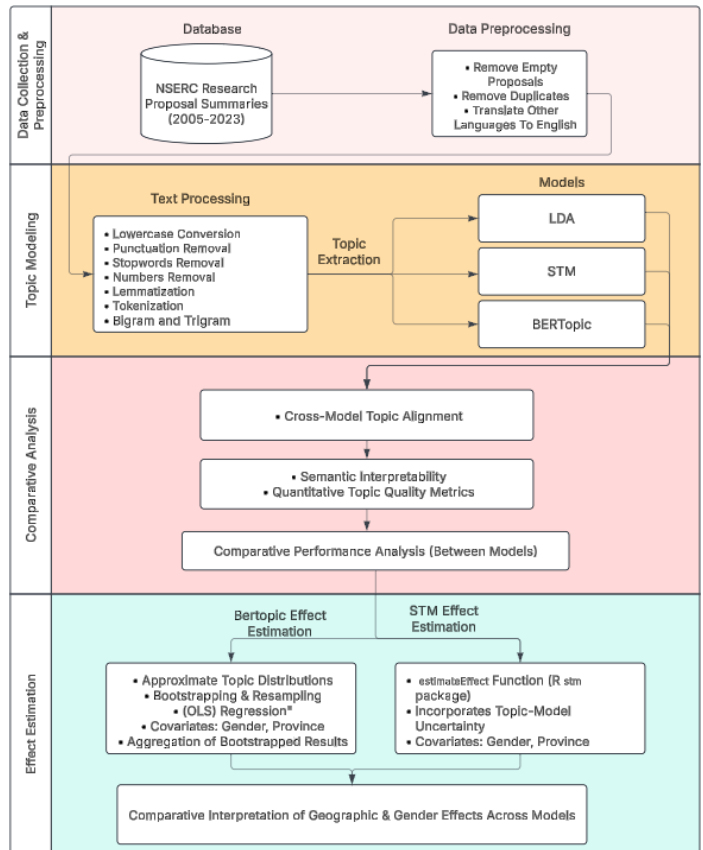

**Figure 2.** High-level flow of the analysis.

## 4.1. Data Preprocessing

The collected data underwent a thorough cleaning and preprocessing process to ensure its compatibility with our modelling techniques. The following steps were implemented:

1. **Removal of incomplete entries:** All entries lacking application summaries were removed to ensure that each record contained sufficient information for analysis.
2. **Elimination of duplicates:** Duplicate records were identified and eliminated to maintain data integrity and prevent redundancy.
3. **Translation of non-English content:** Non-English content, such as French and Italian, was translated into English using the *deep_translator* package in Python. To verify the accuracy of the translations, a random sample of 5% of the translated content was manually reviewed, confirming its suitability for subsequent analysis.
4. **Cleaning of non-textual elements:** Non-textual elements, including punctuation, numbers, and special characters, were removed to streamline the text for processing.
5. **Text normalization:** The text was converted to lowercase and tokenized, facilitating consistent analysis across all entries.
6. **Stop-word removal:** Common English stop-words were removed to reduce noise and focus on meaningful content.
7. **Exclusion of domain-specific terms:** In addition to general stop-words, domain-specific terms such as "Canada", "NSERC", and "Research" were excluded to prevent skewing of thematic analysis.
8. **Lemmatization:** Words were reduced to their root forms by lemmatization, aiding in the consolidation of similar terms.
9. **N-grams:** Bigrams and trigrams were also considered to capture contextually relevant phrases and enhance thematic richness.

This comprehensive preprocessing approach ensured that the dataset was optimized for subsequent modelling and analysis, allowing for accurate exploration of research themes.

## 4.2. Topic Modelling

In this study, we employed and compared three topic modelling techniques to extract themes from our corpus. Each model is described below.

### 4.2.1. Latent Dirichlet Allocation (LDA)

Introduced by Blei et al. (Blei 2003), LDA is a foundational generative probabilistic model designed to uncover latent topics within a document collection. LDA operates under the assumption that each document is a mixture of multiple topics, and each topic is characterized by a distribution of words. To determine the optimal number of topics (K), we conducted a thorough analysis by combining coherence scores ($C_v$) with a grid search spanning 2 to 20 topics. Additionally, we examined the top keywords for each topic to ensure meaningful categorization. Through this approach, we identified 11 as the optimal number of topics.

### 4.2.2. Structural Topic Modelling (STM)

STM (Roberts et al. 2013) enhances the traditional topic modelling framework by integrating document-level metadata, allowing for a nuanced analysis of topic prevalence and content. The STM model was implemented using the *stm* package in R. To determine the optimal number of topics, we employed the *searchK()* function from the same package, conducting a grid search over 2 to 20 topics. This method also identified 11 as the optimal number of topics (K).

### 4.2.3. BERTopic

We employed the BERTopic model (Grootendorst 2022), which effectively combines BERT embeddings with a clustering technique. Initially, document embeddings were generated using a pre-trained BERT model to capture the contextual semantics of each document. To manage the high dimensionality inherent in these embeddings, we applied Uniform Manifold Approximation and Projection (UMAP) (Healy and McInnes 2024) for dimensionality reduction. Subsequently, we employed Hierarchical Density-Based Spatial Clustering of Applications with Noise (HDBSCAN) (McInnes et al. 2017) to cluster similar document embeddings, with each cluster representing a distinct topic. Representative keywords for each topic were extracted using class-based TF-IDF (c-TF-IDF). The resulting topic model was evaluated for coherence and qualitatively assessed to ensure interpretability and relevance. For BERTopic, the optimal number of topics (K) was determined to be 13.

## 4.3. Comparative Analysis of Topic Models

To conduct a robust comparison of the topics generated by the three models—LDA, STM, and BERTopic—we adopted a systematic approach combining cross-model topic alignment, semantic interpretability assessment, and quantitative evaluation metrics.

### 4.3.1. Cross-Model Topic Alignment

To represent each topic consistently, we derived vector representations from the top keywords of each topic. Specifically, we computed the mean vector of Sentence-BERT embeddings (Reimers and Gurevych 2019) for the top-30 keywords of each topic using HuggingFace's *all-MiniLM-L6-v2* pre-trained model. This enabled us to quantify thematic correspondence and overlap among different topic sets by measuring the semantic similarity between their vector representations. We calculated the semantic similarity between any two topic vectors, $v_i$ and $v_j$, using the cosine similarity metric.

Based on these pairwise cosine similarity scores, topics identified by different models were grouped to pinpoint thematic correspondences. Through an iterative qualitative assessment process, we set a grouping threshold at 0.82. We experimented with various thresholds (e.g., 0.70, 0.75, 0.80, 0.85, 0.90), manually inspecting grouped topics for coherence. A lower threshold tended to group dissimilar topics, introducing noise, while a higher one missed clear thematic overlap. The 0.82 threshold emerged as the optimal balance, maximizing meaningful thematic correspondences while minimizing unrelated topic inclusion. Once grouped, these topic groups were categorized as follows:

- **Triplet Matches (n=5):** Groups consisting of one topic from each model (BERTopic, STM, LDA), where the cosine similarity between all three pairs met or exceeded the threshold.
- **Semi-Matches (n=6):** Groups consisting of two topics from different models with a cosine similarity equal to or greater than the threshold, which were not part of a "Triplet Match".

- **Unique Topics (n=8):** Topics from a single model that did not achieve a cosine similarity above the threshold with any topic from another model.

Additionally, a t-SNE (t-Distributed Stochastic Neighbour Embedding) visualization was generated using the Sentence-BERT topic embeddings (van der Maaten and Hinton 2008). This visualization provided a lower-dimensional spatial representation of the topic space, facilitating a qualitative assessment of clustering and interrelationships between topics across different models. To assign thematic labels to each group, we utilized the GPT-4 model (OpenAI Team et al. 2024) to synthesize the core concept from the constituent keywords. Each generated label was then subjected to a final qualitative validation to ensure its fidelity to the underlying concepts of the keyword cluster.

### 4.3.2. Topic Quality Evaluation

For a quantitative comparison of topic quality, we computed three metrics for topics that formed triplet matches. These metrics were calculated using the original lemmatized corpus and the derived dictionary:

#### 4.3.2.1. Topic Coherence ($C_v$)

This metric measures the semantic consistency within a topic based on word co-occurrence patterns in the corpus (Röder et al. 2015). A higher $C_v$ score indicates greater interpretability and reliability. Formally, for a topic T with top words $W_T$, coherence is computed as follows:

$$C_v(T) = \frac{1}{\binom{|W_T|}{2}} \sum_{\substack{W_i, W_j \in W_T \\ i<j}} NPMI(W_i, W_j) \qquad (1)$$

where the average pairwise Normalized Pointwise Mutual Information (NPMI) score is:

$$NPMI(W_i, W_j) = \frac{\log \frac{P(W_i, W_j) + \varepsilon}{P(W_i)P(W_j)}}{-\log(P(W_i, W_j) + \varepsilon)} \qquad (2)$$

with $P(w_i, w_j)$ and $P(w_i)$ estimated from the corpus using co-occurrence counts, and $\varepsilon$ is a small constant to ensure numerical stability.

#### 4.3.2.2. Topic Uniqueness

This metric quantifies the distinctiveness of a topic's top words within the set of evaluated triplet topics. It is calculated as the average inverse frequency of each word among all top words across the triplet topics for a given model. A higher score indicates less word overlap and greater distinctiveness. For a topic T with top words $W_T$:

$$Uniqueness(T) = \frac{1}{|W_T|} \sum_{w \in W_T} \frac{1}{Count(w, W_{all})} \qquad (3)$$

where W_all is the multiset of all top words from the triplet topics for the model, and count(w, W_all) denotes the number of occurrences of word w in W_all. The model's uniqueness score is the average of Uniqueness(T) across all its matched topics.

### 4.3.2.3. Topic Diversity

This metric represents the vocabulary range across the evaluated topics for a given model. It is calculated as the proportion of unique words among all top words from the triplet topics. Formally:

$$Diversity = \frac{|\bigcup_{T \in M} W_T|}{\sum_{T \in M} |W_T|} \quad (4)$$

where M is the set of matched triplet topics for the model, and $\sum_{T \in M} |W_T|$ is the total number of words (with duplicates) across all topics. Higher scores indicate a broader range of distinct terms across topics.

## 4.4. Covariate Effect Estimation

After extracting topics, we aimed to estimate the effects of geographic location and gender on research themes. To address the uneven distribution of scientific publications across provinces, we preprocessed the geographic location variable. Provinces with fewer than 1000 publications were combined into a single ``Other'' group. This step enhances the statistical stability of the regression estimates for the larger, more frequently sampled provinces while still accounting for the contributions from smaller regions.

We estimated the effect of province and gender on the prevalence of each topic identified by the STM and BERTopic model. Topic prevalence, defined as the proportion of a document d assigned to a given topic k ($\theta_{d,k}$), was treated as the dependent variable in a regression model, with province and gender as key predictors separately. The resulting categorical variables were handled using sum contrasts. Sum contrasts compare the mean of each category to the overall mean of the dependent variable across all observations. In this context, the intercept of the regression represents the estimated overall mean topic proportion across all publications, and the coefficient for each province or gender category represents the estimated difference between that category's mean topic proportion and this overall mean. The general form of the regression model for topic k is

$$\theta_{d,k} = \beta_{0,k} + \sum_{j=1}^{P-1} \beta_{j,k} C_{d,j} + \varepsilon_{d,k} \quad (5)$$

where $\theta_{d,k}$ is the proportion of document d in topic k, $\beta_{0,k}$ is the intercept (overall mean), $\beta_{j,k}$ is the coefficient for the j-th category (representing the difference from the overall mean), $C_{d,j}$ is the sum contrast variables for the P categories of the relevant covariate (e.g., gender or geographic location), and $\varepsilon_{d,k}$ is the error term.

For the STM model, we utilized the *estimateEffect* function from the *stm* R package. This function is specifically designed to estimate the effects of document metadata on topic prevalence, incorporating uncertainty by drawing samples from the model's variational posterior distribution. This method provides a principled, model-integrated approach to inference within the STM framework.

For the BERTopic model, which lacks a native probabilistic structure and does not support the same model-integrated inference as STM, we developed an algorithm, named Covariate Effect Estimation for BERTopic (COFFEE). This algorithm conducts a comparable statistical analysis and quantifies uncertainty. Unlike the STM approach, which samples from a model-specific posterior, COFFEE employs bootstrapping, a general resampling technique, to estimate effects. The COFFEE algorithm was developed in Python, providing a robust alternative for effect estimation in the BERTopic framework, as outlined in Algorithm 1.

---

**Algorithm 1** Covariate Effect Estimation for the BERTopic model (COFFEE)

**Require:** Documents $D = \{d_1, \ldots, d_n\}$, pre-trained BERTopic model $\mathcal{M}$, covariate data $C_D$ (e.g., a DataFrame containing Gender or Geographical Location), number of bootstrap samples $N = 25$

**Ensure:** Estimated coefficients $\hat{\beta}_k$, standard errors $SE(\hat{\beta}_k)$, $t$-values $t_k$, and $p$-values $p_k$ for each topic $k$ and each covariate term

1: **Phase 1: Bootstrapped Data Preparation**
2: Initialize `all_theta_samples` and `all_covariate_data_samples` as empty lists
3: **for** $s = 1$ to $N$ **do**
4:     Resample $D^{(s)}$ and $C_D^{(s)}$ with replacement from $D$ and $C_D$
5:     $\Theta^{(s)} \leftarrow \mathcal{M}.\text{approximate\_distribution}(D^{(s)})$
6:     Append $\Theta^{(s)}$ to `all_theta_samples`
7:     Append $C_D^{(s)}$ to `all_covariate_data_samples`
8: **end for**
9: **Phase 2: Per-Sample Regression Analysis**
10: **for** each topic $k$ **do**
11:     Initialize `coef_samples[k]` and `df_resid_samples[k]` as empty lists
12:     **for** $s = 1$ to $N$ **do**
13:         $y^{(s)} \leftarrow$ column $k$ of `all_theta_samples[s]`
14:         $X^{(s)} \leftarrow$ dmatrix(CovariateFormula, data = `all_covariate_data_samples[s]`)
15:         **if** OLS fit is feasible **then**
16:             Fit OLS: $y^{(s)} \sim X^{(s)}$
17:             Append coefficients to `coef_samples[k]`
18:             Append degrees of freedom to `df_resid_samples[k]`
19:         **end if**
20:     **end for**
21: **end for**
22: **Phase 3: Aggregated Inference and Uncertainty Quantification**
23: **for** each topic $k$ **do**
24:     **if** `coef_samples[k]` is not empty **then**
25:         $\hat{\beta}_k \leftarrow$ mean of `coef_samples[k]`
26:         $SE(\hat{\beta}_k) \leftarrow$ standard deviation of `coef_samples[k]`
27:         $df_k \leftarrow$ median of `df_resid_samples[k]`
28:         $t_k \leftarrow \hat{\beta}_k / SE(\hat{\beta}_k)$
29:         $p_k \leftarrow 2 \cdot \text{t.sf}(|t_k|, df_k)$
30:     **else**
31:         $\hat{\beta}_k, SE(\hat{\beta}_k), t_k, p_k \leftarrow$ NaN
32:     **end if**
33: **end for**
34: **return** All computed $\hat{\beta}_k$, $SE(\hat{\beta}_k)$, $t_k$, and $p_k$ for each topic $k$

The COFFEE algorithm enables us to replicate the statistical objective of R's *estimateEffect* function, which regresses topic proportions on metadata using sum contrast. Additionally, COFFEE provides a non-parametric means of approximating the sampling distribution of coefficient estimates through resampling, thereby quantifying uncertainty in the absence of a model-based posterior. By mirroring the analytical structure of STM's *estimateEffect*, this approach allows for a direct and methodologically consistent comparison of the estimated relationships between the selected covariates (e.g., province or gender) and topic prevalence across both STM and BERTopic outputs.

## 5. Results

### 5.1. Comparative Analysis of Topic Models

The comparative analysis highlights both similarities and differences in the thematic structures identified by BERTopic, STM, and LDA, shedding light on each model's strengths and weaknesses. Figure 3 presents the t-SNE visualization of the topics identified by these models. In the plot, clustered points represent semantically similar topics, regardless of their originating model, while more isolated points indicate distinct thematic spaces. This visualization not only confirms quantitative similarities and differences but also serves as a foundational reference for understanding the thematic relationships explored in detail below.

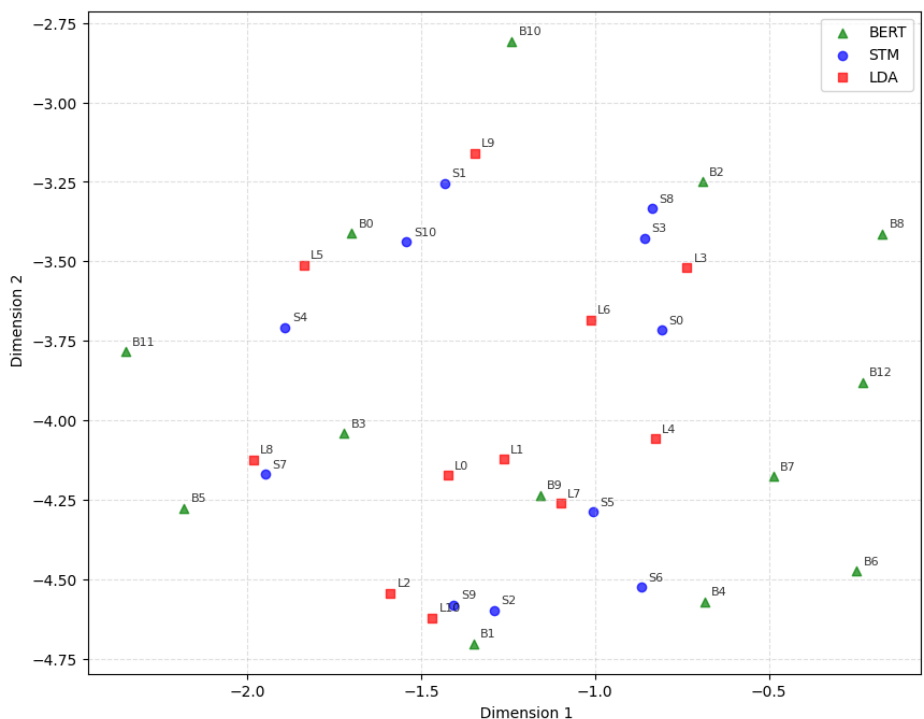

**Figure 3.** t-SNE plot of topics identified by BERTopic, STM, and LDA. Each point represents a topic, and clusters indicate semantic similarity.

The clear groupings in the plot correspond to the triple alignments, where BERTopic, STM, and LDA often cluster together. For example, B0, S10, and L5 are grouped near the top-left, representing "Environmental Science & Industrial Processes", while B1, S2, and L10 are clustered in the lower-right for "Computer Science & Artificial Intelligence". Conversely, the dispersion of unique topics further from these clusters (e.g., B8, B10, B11, B12 at the periphery) visually indicates their distinct semantic spaces and highlights BERTopic's tendency to identify more niche and specialized themes. In this section, we delve deeper into these topic structures, providing a detailed exploration of the thematic alignments and distinctions that characterize each model's outputs.

### 5.1.1. Unique Topics

Table 1 presents a detailed comparison of unique topics identified by each model, highlighting themes predominantly discerned by a single model. These unique topics are visually corroborated by Figure 3, where they often appear as more isolated points, signifying their semantic distinction from the main clusters. Notably, BERTopic stands out with its ability to generate a greater number of specific and unique topics, illustrating its strength in providing a richer semantic understanding of the corpus.

**Table 1.** Comparison of unique topics across models.

| Label | BERTopic | STM | LDA |
|---|---|---|---|
| Musculoskeletal Biomechanics | (6) bone, injury, joint, knee, tissue | --- | --- |
| Dental & Oral Bio-materials | (7) dental, implant, tooth, bone, enamel | --- | --- |
| Public Health & Vaccine Communication | (8) vaccine, vaccine confidence, education, mental health, kmaw | --- | --- |
| Food Science & Agricultural Products | (10) flaxseed, food, fatty acid, mustard, ingredient | --- | --- |
| Advanced Energy Storage & Electronics | (11) battery, energy storage, ion battery, solar cell, transistor | --- | --- |
| Organ Transplantation & Biofabrication | (12) organ, kidney, transplant, donor, blood | --- | --- |
| Plant Pathology & Crop Genetics | --- | (3) disease, plant, dna, gene, bacterium | --- |
| Theoretical & Computational Sciences | --- | (9) theory, flow, dynamic, simulation, computational | --- |

**Note:** Numbers in parentheses represent the topic index.

BERTopic's distinctive strength is exemplified by its unique topic "Public Health & Vaccine Communication" (B8) topic. The top keywords—such as {vaccine, vaccine confidence, education, mental health, kmaw, mi kmaw, covid, 19, covid 19, social medium, indigenous}— illustrate BERTopic's capacity to capture a highly specific and timely theme. This theme encompasses not only medical terms but also socio-political dimensions, such as "vaccine confidence" and culturally specific references ("kmaw", "indigenous"). This level of fine-grained resolution is achieved through BERTopic's advanced contextual embeddings, which are adept at identifying multi-word expressions and emerging research areas that other models might overlook due to their reliance on less context-aware word representations.

Similarly, BERTopic's identification of the "Organ Transplantation & Biofabrication" (B12) topic, with keywords such as {organ, kidney, transplant, transplantation, donor}, and other key terms such as "bioink" and "bioprinting", underscores its ability to isolate specialized subfields within medicine. This capability to delve into niche scientific domains highlights BERTopic's enhanced sensitivity to context and specificity. In contrast, STM also generated meaningful unique topics, such as "Plant Pathology & Crop Genetics" (S3), characterized by keywords such as {disease, plant, crop, dna, gene, bacterium}. However, these topics often exhibited a broader scope compared to BERTopic's highly specialized themes. This may suggest that while STM is effective at identifying significant research areas, it may not capture the same level of detail in emerging or specialized topics as BERTopic. LDA did not produce any unique topics. All LDA topics had at least a semi-match, i.e., partial thematic alignment, with those from BERTopic or STM. This pattern suggests that LDA's reliance on word co-occurrence may lead to the identification of more generalized topics, often blending distinct concepts rather than distinguishing them. Consequently, LDA provides a less detailed thematic landscape compared to the more context-aware capabilities of BERTopic.

### 5.1.2. Triple Alignment

Five topics were consistently identified by all three topic models—BERTopic, STM, and LDA—demonstrating a robust alignment among these different models (see Table 2). This consensus, underscored by an average cosine similarity of 0.935, validates the inherent prominence of these research themes within the dataset, irrespective of the model used. While the models agreed on the core subjects, an in-depth analysis of their top keywords uncovers subtle differences in emphasis and granularity, highlighting the distinct interpretive lens each model applies.

**Table 2.** Consensus topics across BERTopic, STM, and LDA.

| Label | BERTopic | STM | LDA | Avg. Sim. |
|---|---|---|---|---|
| Environmental Science & Industrial Processes | (0) water, energy, climate, gas, soil | (10) water, production, oil, treatment, gas | (5) chemical, reaction, gas, organic, compound | 0.935 |
| Computer Science & Artificial Intelligence | (1) network, algorithm, software, communication, wireless | (2) network, software, technology, algorithm, communication | (10) network, computer, software, theory, algorithm | 0.925 |
| Neuroscience & Cognitive Science | (4) brain, memory, neuron, neural, mechanism | (6) learn, brain, memory, neural, visual | (4) cell, tissue, brain, signal, memory | 0.915 |
| Molecular Biology & Biotechnology | (2) protein, gene, plant, genetic, disease | (0) cell, protein, mechanism, molecular, role | (6) protein, molecular, mechanism, molecule, biological | 0.892 |
| Materials Science & Applied Physics (Imaging & Photonics) | (3) polymer, imaging, laser, optical, molecular | (7) quantum, material, light, optical, particle | (2) image, optical, sensor, measurement, light | 0.867 |

**Note:** Numbers in parentheses represent the topic index.

For the "Environmental Science & Industrial Processes" domain, all three models delineated a coherent topic centred on natural resources, energy, and industrial operations. BERTopic's B0 topic connected industrial activities with broader environmental concerns, using keywords such as {water, energy, climate, gas, soil, technology, industry, carbon, oil, production}. In contrast, LDA's L5 topic focused more on chemical processes, with keywords such as {chemical, reaction, gas, production, organic, compound, metal, water, fuel, produce}, offering a detailed view of the industry's material aspects. STM's S10, with keywords such as {water, production, oil, treatment, company, gas, quality, chemical, industry, environmental}, provided a balanced perspective, integrating industrial operations and resource management, highlighting "treatment" and "quality".

In the "Computer Science & Artificial Intelligence" theme, BERTopic's B1 swiftly captured foundational computing terms, also indicating AI paradigms with keywords such as "machine" and "learning". This suggests a more contemporary focus. LDA's L10, however, remained centred on core elements of computer science without immediate AI cues. STM's S2 presented a solid applied perspective, focusing on practical applications and industry challenges. The "Molecular Biology & Biotechnology" topic showed remarkable semantic overlap across models. BERTopic's B2, STM's S0, and LDA's L7 all yielded highly coherent and relevant terms defining this scientific domain. This alignment indicates a robust thematic signal for well-established scientific fields, demonstrating the models' ability to capture widely recognized domains effectively.

The quantitative evaluation of topic quality, summarized in Table 3, further highlights the comparative performance of the models. BERTopic demonstrated the highest average coherence ($C_V$) at 0.638, outperforming STM (0.604) and LDA (0.569). This suggests BERTopic's embedding approach more effectively groups semantically similar words, resulting in more interpretable and cohesive topics. All models showed high average topic uniqueness, with BERTopic slightly leading at 0.963, indicating distinct topic definitions due to minimal word repetition across topics. While LDA and STM excelled in topic diversity (0.960 and 0.959, respectively), BERTopic had slightly lower diversity (0.953), reflecting its focus on generating more unique, specialized topics rather than using a broad vocabulary across common themes.

**Table 3.** Quantitative evaluation of topic quality.

| Model | Average Coherence ($C_V$) | Average Uniqueness | Average Diversity |
|---|---|---|---|
| BERTopic | 0.638 | 0.963 | 0.953 |
| STM | 0.604 | 0.959 | 0.959 |
| LDA | 0.569 | 0.960 | 0.960 |

### 5.1.3. Partial Alignment

The six partially aligned topics (see Table 4) highlight instances where two models identified similar themes, while the third either did not resolve a comparable topic or subsumed its component words into a broader theme. Interestingly, all observed partial alignments occurred between STM and LDA, or BERTopic and LDA, with no strong pairings between BERTopic and STM reaching the defined similarity threshold. This pattern may indicate that while BERTopic shares some thematic boundaries with LDA, and STM with LDA, the direct semantic alignment between BERTopic and STM's specific topic groupings is less pronounced. This divergence likely stems from their distinct algorithmic approaches to word and document representation.

**Table 4.** Partially aligned topics across BERTopic, STM, and LDA.

| Label | BERTopic | STM | LDA | Avg. Sim. |
|---|---|---|---|---|
| Climate & Aquatic Ecology | --- | (1) climate, water, forest, ecosystem, soil | (9) water, climate, soil, fish, ecosystem | 0.941 |
| Imaging & Sensing Applications | --- | (5) image, sensor, detection, patient, medical | (7) environment, flow, rate, measurement, region | 0.890 |
| Energy & Advanced Materials | --- | (4) energy, material, fuel, polymer, heat | (0) material, energy, industry, polymer, operation | 0.884 |
| Population & Evolutionary Biology | --- | (8) species, population, plant, evolution, fish | (3) gene, population, evolution, specie, gene expression | 0.868 |
| Mechanical Eng. & Sports Technology | (9) bike, composite, sport, suspension, ride | --- | (1) mechanical, structural, vibration, weight, force | 0.852 |
| Quantum & Nuclear Physics | (5) quantum, star, galaxy, matter, physics | --- | (8) nuclear, chip, microfluidic, radiation, spectroscopy | 0.823 |

**Note:** Numbers in parentheses represent the topic index.

For example, the strong alignment between LDA's L9 and STM's S1 topics on "Climate & Aquatic Ecology" demonstrated a shared understanding of environmental and ecosystem assessment, with both models providing relevant terms such as {water, carbon, forest, climate, ecosystem}. In another instance, BERTopic's B9 and LDA's L1 topics showed significant agreement on "Mechanical Engineering & Sports Technology", with BERTopic's keywords such as {design, system, control, structure, vehicle, model, sensor, movement, frame} closely matching LDA's focus on {system, control, design, vehicle, structure, model, data, engineering, movement}.

These observed differences between models can be attributed to their underlying algorithms:

- **BERTopic:** Leverages pre-trained transformer models to generate context-aware word embeddings. This approach allows BERTopic to capture nuanced semantic relationships and identify more granular, specific, and often novel topics, as evidenced by its higher number of highly specialized, unique topics. Its high coherence suggests its ability to form semantically tightly-knit and interpretable clusters of words. This makes BERTopic particularly insightful for uncovering fine-grained thematic structures and emerging areas within a dataset.
- **STM:** A statistical topic model that can incorporate metadata, STM's topics often represent coherent semantic units and demonstrate good overlap with both BERTopic and LDA for broad themes. STM's unique topics are also interpretable, showing a robust capability to identify distinct themes, though generally at a slightly broader level of granularity than BERTopic's most specialized topics.

- **LDA:** As a generative probabilistic model, LDA assumes topics are distributions over words, and documents are distributions over topics. LDA excels at identifying common themes through word co-occurrence patterns, often producing broader, more generalized topics. In this analysis, all LDA topics had at least partially aligned topics with other models, demonstrating strong topic diversity for these broader themes. However, LDA's main comparative weakness lies in its tendency to produce more generalized topics and its inability to resolve the highly specific or novel topics that BERTopic consistently uncovers. This often results in less granular and potentially less actionable insights for specialized research.

### 5.2. Covariate Effect Estimation

In addition to the comparative analysis, a key contribution of this work is the development of the COFFEE algorithm, which allows for robust statistical analysis and effect estimation from BERTopic's non-probabilistic outputs. By systematically comparing the insights derived from BERTopic, processed via COFFEE (see Tables 12 and 14 in Appendix A) against those obtained from Structural Topic Model (STM) (see Tables 13 and 15 in Appendix A, for more details), we highlight BERTopic's ability to corroborate established patterns while revealing unique, granular themes that traditional models such as STM may overlook.

#### 5.2.1. Geographical Location

BERTopic's analysis reveals distinct regional research specializations, offering insights into the contributions of different Canadian provinces. For the topic of "Environmental Science & Industrial Processes" (Table 5), BERTopic highlights a significant positive effect in Alberta (Estimate: 0.0229, $p < 0.0001$). This finding is strongly corroborated by STM's corresponding "Environmental Science & Industrial Processes" topic (Estimate: 0.0202, $p < 0.0001$), aligning with Alberta's well-established prominence in energy and environmental research (Dubé et al. 2022). Moreover, BERTopic also identifies a statistically significant positive effect for this topic in Newfoundland and Labrador (Estimate: 0.0128, $p = 0.0224$), attributed to its unique industrial and environmental context (Gray 2005). STM, however, does not detect a significant effect in this province (Estimate: 0.0070, $p = 0.1952$).

**Table 5.** BERTopic provincial effects, environmental science and industrial processes topic.

| Province | Estimate | Std. Error | t-value | p-value |
|---|---|---|---|---|
| Intercept | 0.0568 | 0.0014 | 39.6711 | <0.0001 |
| Alberta | 0.0229 | 0.0028 | 8.1083 | <0.0001 |
| British Columbia | 0.0062 | 0.0022 | 2.8072 | 0.0050 |
| Manitoba | −0.0003 | 0.0038 | −0.0818 | 0.9348 |
| New Brunswick | −0.0027 | 0.0057 | −0.4749 | 0.6348 |
| Newfoundland and Labrador | 0.0128 | 0.0056 | 2.2831 | 0.0224 |
| Nova Scotia | 0.0006 | 0.0039 | 0.1566 | 0.8756 |
| Ontario | −0.0042 | 0.0021 | −2.0046 | 0.0450 |
| Other | −0.0477 | 0.0021 | −22.7762 | <0.0001 |
| Quebec | −0.0002 | 0.0021 | −0.0934 | 0.9255 |

In the domain of "Computer Science & Artificial Intelligence" (Table 6), both BERTopic and STM confirm strong positive effects in Ontario (BERTopic: 0.0295, p < 0.0001; STM: 0.0239, p < 0.0001) and British Columbia (BERTopic: 0.0051, p = 0.0256; STM: 0.0120, p < 0.0001), reflecting these provinces' role as major technology and AI research hubs (Arnaout et al. 2024; Attard-Frost, Brandusescu, and Lyons 2024). However, BERTopic detects a significant negative effect in Nova Scotia (Estimate: -0.0103, p = 0.0024), a nuance missed by STM (Estimate: -0.0074, p = 0.0807). This might indicate BERTopic's enhanced sensitivity to regional variations and its ability to capture lower concentrations of high-impact research in less tech-centric areas.

**Table 6.** BERTopic provincial effects, computer science and artificial intelligence topic.

| Province | Estimate | Std. Error | t-value | p-value |
| --- | --- | --- | --- | --- |
| Intercept | 0.0449 | 0.0015 | 30.8181 | <0.0001 |
| Alberta | 0.0003 | 0.0024 | 0.1369 | 0.8911 |
| British Columbia | 0.0051 | 0.0023 | 2.2319 | 0.0256 |
| Manitoba | 0.0011 | 0.0037 | 0.3078 | 0.7583 |
| New Brunswick | 0.0045 | 0.0066 | 0.6816 | 0.4955 |
| Newfoundland and Labrador | 0.0017 | 0.0051 | 0.3329 | 0.7392 |
| Nova Scotia | −0.0103 | 0.0034 | −3.0367 | 0.0024 |
| Ontario | 0.0295 | 0.0018 | 5.1528 | <0.0001 |
| Other | −0.0151 | 0.0030 | −5.0668 | <0.0001 |
| Quebec | −0.0046 | 0.0015 | −3.1233 | 0.0018 |

For "Molecular Biology & Biotechnology" (Table 7), both models highlight Manitoba's prominence (BERTopic: 0.0206, p < 0.0001; STM: 0.0230, p < 0.0001), affirming its significant research contributions in life sciences and biotechnology.

**Table 7.** BERTopic provincial effects, molecular biology and biotechnology topic.

| Province | Estimate | Std. Error | t-value | p-value |
| --- | --- | --- | --- | --- |
| Intercept | 0.0586 | 0.0010 | 56.2270 | <0.0001 |
| Alberta | −0.0063 | 0.0018 | −3.4805 | 0.0005 |
| British Columbia | −0.0019 | 0.0024 | −0.7730 | 0.4395 |
| Manitoba | 0.0206 | 0.0042 | 4.8793 | <0.0001 |
| New Brunswick | 0.0140 | 0.0062 | 2.2621 | 0.0237 |
| Newfoundland and Labrador | −0.0056 | 0.0048 | −1.1761 | 0.2396 |
| Nova Scotia | 0.0069 | 0.0031 | 2.2466 | 0.0247 |
| Ontario | −0.0083 | 0.0015 | −5.3439 | <0.0001 |
| Other | −0.0343 | 0.0034 | −10.0927 | <0.0001 |
| Quebec | −0.0097 | 0.0015 | −6.2402 | <0.0001 |

In "Materials Science & Applied Physics" (Table 8), BERTopic uniquely identifies a significant positive effect in New Brunswick (Estimate: 0.0181, p < 0.0001). STM's closest equivalent, "Quantum Physics", does not capture this effect (Estimate: -0.0021, p = 0.6952).

**Table 8.** BERTopic provincial effects, materials science and applied physics topic.

| Province | Estimate | Std. Error | t-value | p-value |
| --- | --- | --- | --- | --- |
| Intercept | 0.0292 | 0.0007 | 40.5076 | <0.0001 |
| Alberta | −0.0007 | 0.0017 | −0.3927 | 0.6945 |
| British Columbia | 0.0015 | 0.0015 | 0.9940 | 0.3202 |
| Manitoba | 0.0042 | 0.0040 | 1.0559 | 0.2910 |
| New Brunswick | 0.0181 | 0.0035 | 5.1810 | <0.0001 |
| Newfoundland and Labrador | 0.0017 | 0.0042 | 0.3983 | 0.6904 |
| Nova Scotia | −0.0005 | 0.0026 | −0.1872 | 0.8515 |
| Ontario | 0.0008 | 0.0012 | 0.6583 | 0.5103 |
| Other | −0.0208 | 0.0015 | −13.7424 | <0.0001 |
| Quebec | −0.0009 | 0.0013 | −0.7276 | 0.4668 |

Finally, in "Public Health & Vaccine Communication" (Table 9), BERTopic identifies significant positive effects in Alberta (Estimate: 0.0042, p = 0.0022), and Ontario (Estimate: 0.0047, p < 0.0001), which are supported by the literature (e.g., (Burney, Donelle, and Kothari 2025; Lang et al. 2021)). STM lacks a direct equivalent topic.

**Table 9.** BERTopic provincial effects, public health and vaccine communication topic.

| Province | Estimate | Std. Error | t-value | p-value |
| --- | --- | --- | --- | --- |
| Intercept | 0.0154 | 0.0007 | 22.1753 | <0.0001 |
| Alberta | 0.0042 | 0.0014 | 3.0594 | 0.0022 |
| British Columbia | 0.0015 | 0.0014 | 1.0831 | 0.2788 |
| Manitoba | 0.0018 | 0.0022 | 0.7915 | 0.4286 |
| New Brunswick | −0.0053 | 0.0024 | −2.1951 | 0.0282 |
| Newfoundland and Labrador | −0.0125 | 0.0013 | −9.4260 | <0.0001 |
| Nova Scotia | 0.0026 | 0.0018 | 1.4691 | 0.1418 |
| Ontario | 0.0047 | 0.0010 | 4.6628 | <0.0001 |
| Other | −0.0061 | 0.0022 | −2.7444 | 0.0061 |
| Quebec | −0.0021 | 0.0010 | −2.1242 | 0.0337 |

Overall, these findings showcase BERTopic's capability, enhanced by the COFFEE algorithm, to capture nuanced regional research patterns with a level of detail and specificity that surpasses traditional methods. This results in a more comprehensive understanding of geographical influences on research themes across Canada.

### 5.2.2. Gender

BERTopic's fine-grained topic resolution, uncovers significant gender-based patterns in research topics. In "Computer Science & Artificial Intelligence" (Table 10), both BERTopic (Estimate: -0.0034, p < 0.0001) and STM (Estimate: -0.0165, p < 0.0001) consistently indicate a stronger association with male researchers. This robust agreement across models aligns with widely documented gender disparities and the underrepresentation of women in STEM fields, particularly in computing and AI \citep{Hango2013GenderDifferences}. The negative estimates for females highlight a gender gap, calling for targeted initiatives to encourage female participation in these critical areas of technological advancement.

**Table 10.** BERTopic gender effects, computer science and artificial intelligence topic.

| Gender | Estimate | Std. Error | t-value | p-value |
|---|---|---|---|---|
| Intercept | 0.0461 | 0.0008 | 58.0180 | <0.0001 |
| Female | −0.0034 | 0.0008 | −4.2460 | <0.0001 |

Similarly, for "Public Health & Vaccine Communication" (Table 11), BERTopic uniquely identifies a significant positive effect for female researchers (Estimate: 0.0029, p < 0.0001). This finding, not captured by STM due to the absence of a direct equivalent topic, underscores the COFFEE algorithm's ability to better detect gender-specific contributions. The positive estimate for females reflects the prominent role of women in public health professions, nursing, and health communication (Canadian Nurses Association 2023).

**Table 11.** BERTopic gender effects, public health and vaccine communication topic.

| Gender | Estimate | Std. Error | t-value | p-value |
|---|---|---|---|---|
| Intercept | 0.0186 | 0.0005 | 35.2883 | <0.0001 |
| Female | 0.0029 | 0.0006 | 5.0659 | <0.0001 |

Overall, these findings illustrate COFFEE-powered BERTopic's nuanced capability to reveal gender-based research trends, offering a deeper understanding of how gender influences thematic contributions in various fields. By highlighting both disparities and areas where women are making significant impacts, this analysis can provide valuable insights that can inform policy decisions and initiatives aimed at promoting gender equity in research.

## 6. Conclusion

This study presents a comprehensive comparative analysis of BERTopic, STM, and LDA, highlighting their distinct capabilities in uncovering thematic structures from a large corpus of Canadian NSERC-funded research proposals within the period of 2005-2023. Our findings demonstrate that while all models effectively identify prominent scientific domains, their thematic resolution capabilities offer unique characteristics crucial for informed methodological selection.

BERTopic consistently exhibited a superior ability to decompose broad subjects into multiple, highly specialized, and semantically rich topics. This enhanced granularity allows for a detailed and nuanced understanding of a domain, as it resolves distinct facets that other models might blend into broader categories. The contextual nature of its embeddings enables BERTopic to grasp subtle semantic relationships, resulting in more insightful and coherent topic definitions and providing uniquely nuanced semantic discovery. STM generally provided a good balance between identifying broad, well-established themes and reasonably distinct sub-topics. Its unique topics are coherent and well-defined, indicating a solid capability to identify meaningful clusters. It serves as a robust option when a moderate level of thematic granularity is desired. Conversely, LDA, while effective at identifying main thematic currents and demonstrating good topic diversity, consistently produced more generalized topics. Its probabilistic nature, relying on word co-occurrence, often results in less specific or insightful topics for dissecting niche research areas, making it less effective for very fine-grained thematic analysis. Collectively, this research offers methodological guidance for selecting appropriate topic modelling approaches based on desired analytical granularity. While all models robustly identify prominent scientific domains, BERTopic excels in uncovering more granular, highly specific, and often novel themes.

Another key contribution of this work is the development and application of the COFFEE algorithm. This novel bootstrap-based pipeline addresses the inferential limitations of non-probabilistic topic models, enabling robust statistical analysis of covariate effects. By pairing COFFEE with BERTopic and comparing the results to STM's established *estimateEffect* function, we validated our approach by corroborating known research specializations. For instance, both frameworks identified Alberta's leadership in "Environmental Science & Industrial Processes" and the prominence of "Computer Science & Artificial Intelligence" in Ontario and British Columbia. Our findings highlighted the superior analytical resolution of the COFFEE-powered BERTopic effect estimate approach. It uncovered unique, statistically significant regional niches that were invisible to STM. For example, BERTopic uniquely detected a significant research focus on "Environmental Science" in Newfoundland and Labrador and on "Materials Science & Applied Physics" in New Brunswick. The analysis of gender effects was equally revealing. While both models confirmed the well-documented underrepresentation of women in "Computer Science & AI", COFFEE uniquely identified "Public Health & Vaccine Communication" as a field with a significant positive association with female researchers. This finding is particularly potent, as STM could not even test this relationship, having failed to identify the topic in the first place.

These findings have significant implications for science policy. By providing a more granular and sensitive analytical tool, the COFFEE-powered BERTopic framework allows funding agencies such as NSERC to move beyond high-level summaries toward a more nuanced understanding of the research landscape. Such precision is vital for developing targeted, evidence-based strategies that support regional research ecosystems and promote the goals of Equity, Diversity, and Inclusion (EDI).

## 7. Limitations and Future Work

While this study provides valuable insights into funded research trends and the performance of various topic models, several limitations should be acknowledged. First, the quality of topic modelling outputs is significantly influenced by the preprocessing step involved. Although we ensured consistency across all models for direct comparison, variations in tokenization, lemmatization, and stop-word removal could affect the result. Future studies could explore the impact of different preprocessing techniques to assess their influence on model outcomes. Second, the choice of the 0.82 cosine similarity threshold for defining topic similarity between models is a hyperparameter. This threshold was determined through an iterative qualitative assessment to optimize meaningful thematic correspondences. However, selecting a different threshold could alter the categorization of topics. Further research could examine the effects of varying this parameter and develop methods for dynamically adjusting it based on dataset characteristics. Third, the interpretation and labelling of topics, initially generated using the GPT-4o model and refined through human review, are inherently subjective. Despite efforts to mitigate bias, different human interpretations and potential biases in large language models should be considered. Developing more objective and automated methods for topic interpretation could enhance reliability. Methodologically, further investigation into robust, model-integrated covariate effect estimation techniques for embedding-based topic models is warranted to reduce reliance on post-hoc bootstrapping. Such advancements would improve the precision and reliability of effect estimation in embedding-based models, e.g., BERTopic. Substantively, future work could expand the dataset to include research proposals funded by other organizations, such as Canadian Tri-Agencies (Canadian Institutes of Health Research (CIHR), Social Sciences and Humanities Research Council of Canada (SSHRC)). This would provide a more comprehensive view of the national research ecosystem. Additionally, a deeper exploration of the underlying factors driving provincial and gender-based specializations, potentially integrating additional demographic variables or historical policy analyses, would offer valuable context. Finally, exploring the temporal evolution of these patterns in greater detail and investigating the relationship between topic prevalence and actual funding outcomes (e.g., success rates, award amounts) could yield critical insights for optimizing science policy and fostering a more inclusive and innovative scientific community.

# Appendix A. Detailed Regresion Results

Table 12: BERTopic Provincial Effects: Regression Coefficients (Estimate, Std. Error, t-value, and P-value)

| BERTopic Topic Name | Province | Estimate | Std. Error | t-value | p-value |
|---|---|---|---|---|---|
| Environmental Science & Industrial Processes | Intercept | 0.0568 | 0.0014 | 39.6711 | <0.0001 |
| | Alberta | 0.0229 | 0.0028 | 8.1083 | <0.0001 |
| | British Columbia | 0.0062 | 0.0022 | 2.8072 | 0.0050 |
| | Manitoba | −0.0003 | 0.0038 | −0.0818 | 0.9348 |
| | New Brunswick | −0.0027 | 0.0057 | −0.4749 | 0.6348 |
| | Newfoundland and Labrador | 0.0128 | 0.0056 | 2.2831 | 0.0224 |
| | Nova Scotia | 0.0006 | 0.0039 | 0.1566 | 0.8756 |
| | Ontario | −0.0042 | 0.0021 | −2.0046 | 0.0450 |
| | Other | −0.0477 | 0.0021 | −22.7762 | <0.0001 |
| | Québec | −0.0002 | 0.0021 | −0.0934 | 0.9255 |
| Computer Science & Artificial Intelligence | Intercept | 0.0449 | 0.0015 | 30.8181 | <0.0001 |
| | Alberta | 0.0003 | 0.0024 | 0.1369 | 0.8911 |
| | British Columbia | 0.0051 | 0.0023 | 2.2319 | 0.0256 |
| | Manitoba | 0.0011 | 0.0037 | 0.3078 | 0.7583 |
| | New Brunswick | 0.0045 | 0.0066 | 0.6816 | 0.4955 |
| | Newfoundland and Labrador | 0.0017 | 0.0051 | 0.3329 | 0.7392 |
| | Nova Scotia | −0.0103 | 0.0034 | −3.0367 | 0.0024 |
| | Ontario | 0.0295 | 0.0018 | 5.1528 | <0.0001 |
| | Other | −0.0151 | 0.0030 | −5.0668 | <0.0001 |
| | Québec | −0.0046 | 0.0015 | −3.1233 | 0.0018 |
| Molecular Biology & Biotechnology | Intercept | 0.0586 | 0.0010 | 56.2270 | <0.0001 |
| | Alberta | −0.0063 | 0.0018 | −3.4805 | 0.0005 |
| | British Columbia | −0.0019 | 0.0024 | −0.7730 | 0.4395 |
| | Manitoba | 0.0206 | 0.0042 | 4.8793 | <0.0001 |
| | New Brunswick | 0.0140 | 0.0062 | 2.2621 | 0.0237 |
| | Newfoundland and Labrador | −0.0056 | 0.0048 | −1.1761 | 0.2396 |
| | Nova Scotia | 0.0069 | 0.0031 | 2.2466 | 0.0247 |
| | Ontario | −0.0083 | 0.0015 | −5.3439 | <0.0001 |
| | Other | −0.0343 | 0.0034 | −10.0927 | <0.0001 |
| | Québec | −0.0097 | 0.0015 | −6.2402 | <0.0001 |
| Materials Science & Applied Physics | | 0.0292 | 0.0007 | 40.5076 | <0.0001 |

| BERTopic Topic Name | Province | Estimate | Std. Error | t-value | p-value |
|---|---|---|---|---|---|
| | Alberta | −0.0007 | 0.0017 | −0.3927 | 0.6945 |
| | British Columbia | 0.0015 | 0.0015 | 0.9940 | 0.3202 |
| | Manitoba | 0.0042 | 0.0040 | 1.0559 | 0.2910 |
| | New Brunswick | 0.0181 | 0.0035 | 5.1810 | <0.0001 |
| | Newfoundland and Labrador | 0.0017 | 0.0042 | 0.3983 | 0.6904 |
| | Nova Scotia | −0.0005 | 0.0026 | −0.1872 | 0.8515 |
| | Ontario | 0.0008 | 0.0012 | 0.6583 | 0.5103 |
| | Other | −0.0208 | 0.0015 | −13.7424 | <0.0001 |
| | Québec | −0.0009 | 0.0013 | −0.7276 | 0.4668 |
| Neuroscience & Cognitive Science | Intercept | 0.0405 | 0.0010 | 40.3336 | <0.0001 |
| | Alberta | 0.0061 | 0.0024 | 2.5869 | 0.0097 |
| | British Columbia | 0.0021 | 0.0020 | 1.0530 | 0.2924 |
| | Manitoba | 0.0086 | 0.0039 | 2.2157 | 0.0267 |
| | New Brunswick | −0.0127 | 0.0042 | −3.0541 | 0.0023 |
| | Newfoundland and Labrador | 0.0165 | 0.0063 | 2.6373 | 0.0084 |
| | Nova Scotia | 0.0039 | 0.0038 | 1.0240 | 0.3058 |
| | Ontario | 0.0124 | 0.0016 | 7.9094 | <0.0001 |
| | Other | −0.0233 | 0.0025 | −9.2669 | <0.0001 |
| | Québec | 0.0038 | 0.0013 | 2.8560 | 0.0043 |
| Quantum & Nuclear Physics | Intercept | 0.0544 | 0.0015 | 36.6840 | <0.0001 |
| | Alberta | −0.0063 | 0.0030 | −2.0952 | 0.0362 |
| | British Columbia | 0.0125 | 0.0025 | 4.9586 | <0.0001 |
| | Manitoba | −0.0006 | 0.0041 | −0.1368 | 0.8912 |
| | New Brunswick | 0.0059 | 0.0063 | 0.9344 | 0.3501 |
| | Newfoundland and Labrador | −0.0113 | 0.0051 | −2.1993 | 0.0279 |
| | Nova Scotia | −0.0093 | 0.0028 | −3.2934 | 0.0010 |
| | Ontario | 0.0056 | 0.0018 | 3.1644 | 0.0016 |
| | Other | 0.0358 | 0.0045 | 7.9018 | <0.0001 |
| | Québec | −0.0153 | 0.0020 | −7.5679 | <0.0001 |
| Musculoskeletal Biomechanics | Intercept | 0.0202 | 0.0009 | 23.2768 | <0.0001 |
| | Alberta | 0.0024 | 0.0015 | 1.6532 | 0.0983 |
| | British Columbia | 0.0018 | 0.0016 | 1.1002 | 0.2712 |
| | Manitoba | 0.0044 | 0.0024 | 1.8389 | 0.0659 |
| | New Brunswick | 0.0055 | 0.0029 | 1.8906 | 0.0587 |

| BERTopic Topic Name | Province | Estimate | Std. Error | t-value | p-value |
|---|---|---|---|---|---|
| | Newfoundland and Labrador | 0.0003 | 0.0035 | 0.0983 | 0.9217 |
| | Nova Scotia | −0.0019 | 0.0023 | −0.8302 | 0.4064 |
| | Ontario | 0.0059 | 0.0010 | 5.7589 | <0.0001 |
| | Other | −0.0157 | 0.0010 | −15.6707 | <0.0001 |
| | Québec | 0.0012 | 0.0011 | 1.1599 | 0.2461 |
| Dental & Oral Biomaterials | Intercept | 0.0074 | 0.0005 | 16.3041 | <0.0001 |
| | Alberta | −0.0003 | 0.0008 | −0.3287 | 0.7424 |
| | British Columbia | −0.0004 | 0.0006 | −0.6092 | 0.5424 |
| | Manitoba | 0.0011 | 0.0014 | 0.8240 | 0.4099 |
| | New Brunswick | −0.0039 | 0.0013 | −2.9155 | 0.0036 |
| | Newfoundland and Labrador | −0.0037 | 0.0014 | −2.5552 | 0.0106 |
| | Nova Scotia | 0.0046 | 0.0018 | 2.5655 | 0.0103 |
| | Ontario | 0.0019 | 0.0007 | 2.7998 | 0.0051 |
| | Other | −0.0046 | 0.0010 | −4.8555 | <0.0001 |
| | Québec | 0.0040 | 0.0008 | 5.2781 | <0.0001 |
| Public Health & Vaccine Communication | Intercept | 0.0154 | 0.0007 | 22.1753 | <0.0001 |
| | Alberta | 0.0042 | 0.0014 | 3.0594 | 0.0022 |
| | British Columbia | 0.0015 | 0.0014 | 1.0831 | 0.2788 |
| | Manitoba | 0.0018 | 0.0022 | 0.7915 | 0.4286 |
| | New Brunswick | −0.0053 | 0.0024 | −2.1951 | 0.0282 |
| | Newfoundland and Labrador | −0.0125 | 0.0013 | −9.4260 | <0.0001 |
| | Nova Scotia | 0.0026 | 0.0018 | 1.4691 | 0.1418 |
| | Ontario | 0.0047 | 0.0010 | 4.6628 | <0.0001 |
| | Other | −0.0061 | 0.0022 | −2.7444 | 0.0061 |
| | Québec | −0.0021 | 0.0010 | −2.1242 | 0.0337 |
| Mechanical Engineering & Sports Technology | Intercept | 0.0043 | 0.0003 | 13.5216 | <0.0001 |
| | Alberta | −0.0001 | 0.0008 | −0.1836 | 0.8544 |
| | British Columbia | 0.0016 | 0.0007 | 2.2140 | 0.0268 |
| | Manitoba | 0.0013 | 0.0016 | 0.8185 | 0.4131 |
| | New Brunswick | 0.0033 | 0.0023 | 1.4372 | 0.1507 |
| | Newfoundland and Labrador | −0.0037 | 0.0005 | −7.3070 | <0.0001 |
| | Nova Scotia | −0.0023 | 0.0008 | −2.9480 | 0.0032 |
| | Ontario | 0.0002 | 0.0004 | 0.3815 | 0.7028 |
| | Other | −0.0043 | 0.0003 | −13.5216 | <0.0001 |

| BERTopic Topic Name | Province | Estimate | Std. Error | t-value | p-value |
|---|---|---|---|---|---|
| | Québec | 0.0060 | 0.0006 | 9.6339 | <0.0001 |
| Food Science & Agricultural Products | Intercept | 0.0607 | 0.0015 | 39.4140 | <0.0001 |
| | Alberta | 0.0341 | 0.0027 | 12.4820 | <0.0001 |
| | British Columbia | −0.0302 | 0.0022 | −13.5390 | <0.0001 |
| | Manitoba | 0.0200 | 0.0043 | 4.6310 | <0.0001 |
| | New Brunswick | −0.0133 | 0.0049 | −2.6965 | 0.0070 |
| | Newfoundland and Labrador | 0.0344 | 0.0073 | 4.7343 | <0.0001 |
| | Nova Scotia | −0.0010 | 0.0047 | −0.2208 | 0.8252 |
| | Ontario | −0.0248 | 0.0017 | −14.2935 | <0.0001 |
| | Other | −0.0311 | 0.0030 | −10.2104 | <0.0001 |
| | Québec | −0.0189 | 0.0022 | −8.7359 | <0.0001 |
| Advanced Energy Storage & Electronics | Intercept | 0.0341 | 0.0010 | 33.1378 | <0.0001 |
| | Alberta | 0.0001 | 0.0015 | 0.0563 | 0.9551 |
| | British Columbia | 0.0034 | 0.0017 | 1.9710 | 0.0487 |
| | Manitoba | −0.0062 | 0.0034 | −1.8373 | 0.0662 |
| | New Brunswick | −0.0076 | 0.0047 | −1.6201 | 0.1052 |
| | Newfoundland and Labrador | −0.0009 | 0.0046 | −0.2022 | 0.8397 |
| | Nova Scotia | 0.0130 | 0.0043 | 2.9984 | 0.0027 |
| | Ontario | 0.0103 | 0.0012 | 8.8871 | <0.0001 |
| | Other | −0.0088 | 0.0028 | −3.0926 | 0.0020 |
| | Québec | 0.0040 | 0.0012 | 3.2651 | 0.0011 |
| Organ Transplantation & Biofabrication | Intercept | 0.0162 | 0.0008 | 21.4995 | <0.0001 |
| | Alberta | 0.0011 | 0.0015 | 0.7286 | 0.4663 |
| | British Columbia | 0.0015 | 0.0008 | 1.8710 | 0.0614 |
| | Manitoba | 0.0042 | 0.0027 | 1.5529 | 0.1204 |
| | New Brunswick | −0.0040 | 0.0030 | −1.3184 | 0.1874 |
| | Newfoundland and Labrador | −0.0005 | 0.0033 | −0.1655 | 0.8686 |
| | Nova Scotia | 0.0002 | 0.0020 | 0.1174 | 0.9065 |
| | Ontario | 0.0050 | 0.0010 | 4.9844 | <0.0001 |
| | Other | −0.0081 | 0.0017 | −4.7325 | <0.0001 |
| | Québec | −0.0002 | 0.0010 | −0.1938 | 0.8463 |

Table 13: STM Provincial Effects: Regression Coefficients (Estimate, Std. Error, t-value, and P-value)

| STM Topic Name | Province | Estimate | Std. Error | t-value | p-value |
|---|---|---|---|---|---|
| Molecular Biology & Biotechnology | Intercept | 0.085398 | 0.001296739 | 65.82200752 | <0.0001 |
| | Alberta | 0.005720 | 0.002587433 | 2.21067115 | 0.0271 |
| | British Columbia | −0.005519 | 0.002311085 | −2.38821320 | 0.0169 |
| | Manitoba | 0.022977 | 0.004678221 | 4.91143808 | <0.0001 |
| | New Brunswick | −0.033149 | 0.005547517 | −5.97548515 | <0.0001 |
| | Newfoundland and Labrador | −0.008513 | 0.005728392 | −1.48611889 | 0.1373 |
| | Nova Scotia | −0.000093 | 0.003844853 | −0.02413360 | 0.9807 |
| | Ontario | 0.003960 | 0.001672512 | 2.36747665 | 0.0179 |
| | Other | 0.006833 | 0.004638616 | 1.47314830 | 0.1407 |
| | Québec | −0.000134 | 0.001978779 | −0.06762522 | 0.9461 |
| Climate & Aquatic Ecology | Intercept | 0.095541 | 0.001172346 | 81.4956768 | <0.0001 |
| | Alberta | −0.010581 | 0.002128638 | −4.9707803 | <0.0001 |
| | British Columbia | −0.002908 | 0.002194286 | −1.3254554 | 0.1850 |
| | Manitoba | 0.000879 | 0.003716402 | 0.2365847 | 0.8130 |
| | New Brunswick | 0.008624 | 0.005171554 | 1.6675706 | 0.0954 |
| | Newfoundland and Labrador | 0.040209 | 0.005859428 | 6.8622488 | <0.0001 |
| | Nova Scotia | 0.019350 | 0.003865613 | 5.0055992 | <0.0001 |
| | Ontario | −0.024309 | 0.001517287 | −16.0211711 | <0.0001 |
| | Other | −0.024444 | 0.003952047 | −6.1850603 | <0.0001 |
| | Québec | −0.016225 | 0.001572214 | −10.3201574 | <0.0001 |
| Imaging & Sensing Applications | Intercept | 0.082008 | 0.0009099623 | 90.122121 | <0.0001 |
| | Alberta | 0.005323 | 0.0018089315 | 2.942798 | 0.0033 |
| | British Columbia | 0.011312 | 0.0016464941 | 6.870585 | <0.0001 |
| | Manitoba | 0.004291 | 0.0030981379 | 1.384946 | 0.1661 |
| | New Brunswick | −0.004191 | 0.0038817043 | −1.079646 | 0.2803 |
| | Newfoundland and Labrador | −0.010432 | 0.0042779196 | −2.438485 | 0.0148 |
| | Nova Scotia | −0.000066 | 0.0027843747 | −0.023580 | 0.9812 |
| | Ontario | 0.012330 | 0.0012148352 | 10.149686 | <0.0001 |
| | Other | −0.009866 | 0.0031072792 | −3.175048 | 0.0015 |
| | Québec | 0.002407 | 0.0013467447 | 1.787240 | 0.0739 |
| Energy & Advanced Materials | Intercept | 0.0341 | 0.0010 | 33.1378 | <0.0001 |
| | Alberta | 0.0001 | 0.0015 | 0.0563 | 0.9551 |
| | British Columbia | 0.0034 | 0.0017 | 1.9710 | 0.0487 |
| | Manitoba | −0.0062 | 0.0034 | −1.8373 | 0.0662 |
| | New Brunswick | −0.0076 | 0.0047 | −1.6201 | 0.1052 |

| STM Topic Name | Province | Estimate | Std. Error | t-value | p-value |
|---|---|---|---|---|---|
| | Newfoundland and Labrador | −0.0009 | 0.0046 | −0.2022 | 0.8397 |
| | Nova Scotia | 0.0130 | 0.0043 | 2.9984 | 0.0027 |
| | Ontario | 0.0103 | 0.0012 | 8.8871 | <0.0001 |
| | Other | −0.0088 | 0.0028 | −3.0926 | 0.0020 |
| | Québec | 0.0040 | 0.0012 | 3.2651 | 0.0011 |
| Population & Evolutionary Biology | Intercept | 0.070165 | 0.0009938449 | 70.599348 | <0.0001 |
| | Alberta | −0.011328 | 0.0018350971 | −6.173099 | <0.0001 |
| | British Columbia | −0.005862 | 0.0016216842 | −3.614810 | 0.0003 |
| | Manitoba | −0.005278 | 0.0030941796 | −1.705921 | 0.0880 |
| | New Brunswick | 0.017968 | 0.0043005931 | 4.177931 | <0.0001 |
| | Newfoundland and Labrador | 0.016385 | 0.0047886124 | 3.421634 | 0.0006 |
| | Nova Scotia | 0.009955 | 0.0028696285 | 3.468938 | 0.0005 |
| | Ontario | −0.014654 | 0.0012886342 | −11.371713 | <0.0001 |
| | Other | 0.008620 | 0.0034894326 | 2.470388 | 0.0135 |
| | Québec | −0.019649 | 0.0014176333 | −13.860443 | <0.0001 |
| Plant Pathology & Crop Genetics | Intercept | 0.060399 | 0.0009284075 | 65.056089 | <0.0001 |
| | Alberta | −0.004180 | 0.0017919540 | −2.332721 | 0.0197 |
| | British Columbia | −0.007875 | 0.0014992006 | −5.252960 | <0.0001 |
| | Manitoba | 0.015327 | 0.0032849578 | 4.665681 | <0.0001 |
| | New Brunswick | −0.007554 | 0.0040259780 | −1.876327 | 0.0606 |
| | Newfoundland and Labrador | −0.009479 | 0.0040869648 | −2.319272 | 0.0204 |
| | Nova Scotia | 0.003975 | 0.0028568817 | 1.391232 | 0.1642 |
| | Ontario | −0.006298 | 0.0011748510 | −5.360685 | <0.0001 |
| | Other | −0.006359 | 0.0031692739 | −2.006498 | 0.0448 |
| | Québec | −0.003394 | 0.0013001334 | −2.610286 | 0.0090 |
| Environmental Science & Industrial Processes | Intercept | 0.099808 | 0.001222136 | 81.667138942 | <0.0001 |
| | Alberta | 0.020189 | 0.002357155 | 8.565108248 | <0.0001 |
| | British Columbia | −0.019502 | 0.001989692 | −9.801467327 | <0.0001 |
| | Manitoba | 0.000014 | 0.004246428 | 0.003208982 | 0.9974 |
| | New Brunswick | −0.008109 | 0.005405837 | −1.500003469 | 0.1336 |
| | Newfoundland and Labrador | 0.007012 | 0.005413351 | 1.295227454 | 0.1952 |
| | Nova Scotia | 0.001286 | 0.003723598 | 0.345346276 | 0.7298 |
| | Ontario | −0.016596 | 0.001542153 | −10.761254657 | <0.0001 |
| | Other | −0.026044 | 0.004014571 | −6.487327314 | <0.0001 |

| STM Topic Name | Province | Estimate | Std. Error | t-value | p-value |
|---|---|---|---|---|---|
| | Québec | 0.018 340 | 0.001 799 288 | 10.193 132 705 | <0.0001 |
| Computer Science & Artificial Intelligence | Intercept | 0.117 127 | 0.001 564 651 | 74.858 055 2 | <0.0001 |
| | Alberta | 0.002 319 | 0.002 797 940 | 0.828 840 3 | 0.4072 |
| | British Columbia | 0.012 043 | 0.002 543 950 | 4.733 944 6 | <0.0001 |
| | Manitoba | −0.010 893 | 0.004 826 980 | −2.256 642 1 | 0.0240 |
| | New Brunswick | 0.009 120 | 0.006 263 021 | 1.456 097 5 | 0.1454 |
| | Newfoundland and Labrador | −0.011 532 | 0.006 838 957 | −1.686 241 5 | 0.0918 |
| | Nova Scotia | −0.007 378 | 0.004 224 324 | −1.746 505 9 | 0.0807 |
| | Ontario | 0.023 915 | 0.001 860 052 | 12.857 403 0 | <0.0001 |
| | Other | −0.026 031 | 0.004 811 495 | −5.410 257 0 | <0.0001 |
| | Québec | 0.015 857 | 0.002 114 645 | 7.498 613 1 | <0.0001 |
| Neuroscience & Cognitive Science | Intercept | 0.075 622 | 0.001 242 872 | 60.844 734 4 | <0.0001 |
| | Alberta | −0.001 777 | 0.002 225 178 | −0.798 597 4 | 0.4245 |
| | British Columbia | 0.003 705 | 0.002 076 270 | 1.784 654 5 | 0.0743 |
| | Manitoba | −0.009 700 | 0.003 857 281 | −2.514 601 9 | 0.0119 |
| | New Brunswick | −0.003 940 | 0.005 047 635 | −0.780 597 7 | 0.4350 |
| | Newfoundland and Labrador | 0.002 314 | 0.005 431 799 | 0.425 997 7 | 0.6701 |
| | Nova Scotia | −0.000 873 | 0.003 479 721 | −0.250 876 1 | 0.8019 |
| | Ontario | 0.010 627 | 0.001 556 865 | 6.826 110 5 | <0.0001 |
| | Other | 0.015 715 | 0.004 037 685 | 3.892 024 0 | <0.0001 |
| | Québec | −0.002 376 | 0.001 636 088 | −1.452 430 1 | 0.1464 |
| Materials Science & Applied Physics (Imaging | Photonics) | 0.087 554 | 0.001 228 597 | 71.2635218 | |
| <0.0001 | | | | | |
| | Alberta | −0.005 900 | 0.002 188 093 | −2.696 276 0 | 0.0070 |
| | British Columbia | 0.008 011 | 0.002 017 129 | 3.971 578 8 | <0.0001 |
| | Manitoba | 0.000 450 | 0.004 201 146 | 0.107 173 6 | 0.9147 |
| | New Brunswick | −0.002 106 | 0.005 376 835 | −0.391 742 4 | 0.6952 |
| | Newfoundland and Labrador | −0.018 065 | 0.005 296 674 | −3.410 595 4 | 0.0006 |
| | Nova Scotia | −0.007 041 | 0.003 562 315 | −1.976 571 8 | 0.0481 |
| | Ontario | −0.000 857 | 0.001 581 161 | −0.541 938 7 | 0.5879 |
| | Other | 0.042 084 | 0.004 007 692 | 10.500 729 2 | <0.0001 |
| | Québec | −0.006 940 | 0.001 763 798 | −3.934 877 4 | <0.0001 |
| Theoretical & Computational Sciences | Intercept | 0.120 172 | 0.001 241 398 | 96.803 670 8 | <0.0001 |

| STM Topic Name | Province | Estimate | Std. Error | t-value | p-value |
|---|---|---|---|---|---|
| | Alberta | −0.006 533 | 0.002 265 466 | −2.883 550 3 | 0.0039 |
| | British Columbia | 0.000 999 | 0.002 092 320 | 0.477 487 2 | 0.6330 |
| | Manitoba | −0.007 488 | 0.004 234 511 | −1.768 226 4 | 0.0770 |
| | New Brunswick | 0.013 590 | 0.005 466 175 | 2.486 143 6 | 0.0129 |
| | Newfoundland and Labrador | 0.017 569 | 0.005 516 460 | 3.184 869 3 | 0.0014 |
| | Nova Scotia | −0.014 568 | 0.003 622 136 | −4.021 953 6 | <0.0001 |
| | Ontario | 0.000 814 | 0.001 558 672 | 0.522 184 8 | 0.6015 |
| | Other | 0.017 909 | 0.004 225 141 | 4.238 779 0 | <0.0001 |
| | Québec | −0.005 235 | 0.001 811 952 | −2.889 232 2 | 0.0039 |

Table 14: BERTopic Gender Effects: Regression Coefficients (Estimate, Std. Error, t-value, and P-value)

| BERTopic Topic Name | Gender Effect | Estimate | Std. Error | t-value | p-value |
|---|---|---|---|---|---|
| Environmental Science & Industrial Processes | Intercept | 0.0565 | 0.0008 | 71.4637 | <0.0001 |
| | Female | −0.0022 | 0.0009 | −2.5313 | 0.0114 |
| Computer Science & Artificial Intelligence | Intercept | 0.0461 | 0.0008 | 58.0180 | <0.0001 |
| | Female | −0.0034 | 0.0008 | −4.2460 | <0.0001 |
| Molecular Biology & Biotechnology | Intercept | 0.0545 | 0.0009 | 61.6490 | <0.0001 |
| | Female | 0.0033 | 0.0009 | 3.5601 | 0.0004 |
| Materials Science & Applied Physics (Imaging | Photonics) | | 0.0274 | 0.0005 | 55.4532 |
| <0.0001 | | | | | |
| | Female | −0.0037 | 0.0005 | −6.9179 | <0.0001 |
| Neuroscience & Cognitive Science | Intercept | 0.0508 | 0.0009 | 55.8000 | <0.0001 |
| | Female | 0.0086 | 0.0008 | 10.7550 | <0.0001 |
| Quantum & Nuclear Physics | Intercept | 0.0486 | 0.0007 | 70.1653 | <0.0001 |
| | Female | −0.0104 | 0.0008 | −13.8659 | <0.0001 |
| Musculoskeletal Biomechanics | Intercept | 0.0234 | 0.0006 | 36.7802 | <0.0001 |
| | Female | 0.0013 | 0.0005 | 2.7189 | 0.0066 |
| Dental & Oral Biomaterials | Intercept | 0.0092 | 0.0003 | 27.2377 | <0.0001 |
| | Female | 0.0003 | 0.0003 | 0.7716 | 0.4403 |
| Public Health & Vaccine Communication | Intercept | 0.0186 | 0.0005 | 35.2883 | <0.0001 |

| BERTopic Topic Name | Gender Effect | Estimate | Std. Error | t-value | p-value |
|---|---|---|---|---|---|
| | Female | 0.0029 | 0.0006 | 5.0659 | <0.0001 |
| Mechanical Engineering & Sports Technology | Intercept | 0.0055 | 0.0003 | 16.8807 | <0.0001 |
| | Female | −0.0006 | 0.0003 | −1.6514 | 0.0987 |
| Food Science & Agricultural Products | Intercept | 0.0476 | 0.0009 | 54.1775 | <0.0001 |
| | Female | 0.0021 | 0.0008 | 2.5745 | 0.0100 |
| Advanced Energy Storage & Electronics | Intercept | 0.0363 | 0.0006 | 59.0322 | <0.0001 |
| | Female | −0.0065 | 0.0006 | −10.3217 | <0.0001 |
| Organ Transplantation & Biofabrication | Intercept | 0.0192 | 0.0005 | 37.9411 | <0.0001 |
| | Female | 0.0019 | 0.0005 | 3.7516 | 0.0002 |

Table 15: STM Gender Effects: Regression Coefficients (Estimate, Std. Error, t-value, and P-value)

| STM Topic Name | Gender Effect | Estimate | Std. Error | t-value | p-value |
|---|---|---|---|---|---|
| Molecular Biology & Biotechnology | Intercept | 0.10187417 | 0.001531854 | 66.50385 | <0.0001 |
| | Female | 0.02000307 | 0.001742254 | 11.48114 | <0.0001 |
| Climate & Aquatic Ecology | Intercept | 0.09265462 | 0.001314964 | 70.461712 | <0.0001 |
| | Female | 0.01481620 | 0.001515709 | 9.775095 | <0.0001 |
| Imaging & Sensing Applications | Intercept | 0.084198294 | 0.001062309 | 79.259684 | <0.0001 |
| | Female | −0.006128096 | 0.001296875 | −4.725279 | 0.00000230 |
| Energy & Advanced Materials | Intercept | 0.09990712 | 0.001469472 | 67.98844 | <0.0001 |
| | Female | −0.02090085 | 0.001684015 | −12.41132 | <0.0001 |
| Population & Evolutionary Biology | Intercept | 0.07303447 | 0.001132230 | 64.50495 | <0.0001 |
| | Female | 0.01920903 | 0.001337477 | 14.36214 | <0.0001 |
| Plant Pathology & Crop Genetics | Intercept | 0.06425817 | 0.001014583 | 63.334588 | <0.0001 |
| | Female | 0.01034527 | 0.001167860 | 8.858313 | <0.0001 |
| Environmental Science & Industrial Processes | Intercept | 0.103305811 | 0.001426416 | 72.423332 | <0.0001 |
| | Female | 0.007440839 | 0.001697746 | 4.382775 | 0.00001173 |
| Computer Science & Artificial Intelligence | Intercept | 0.11862093 | 0.001503696 | 78.886250 | <0.0001 |
| | Female | −0.01649252 | 0.001776533 | −9.283544 | <0.0001 |

| STM Topic Name | Gender Effect | Estimate | Std. Error | t-value | p-value |
| --- | --- | --- | --- | --- | --- |
| Neuroscience & Cognitive Science | Intercept | 0.09168592 | 0.001346638 | 68.08505 | <0.0001 |
| | Female | 0.01701871 | 0.001599775 | 10.63819 | <0.0001 |
| Materials Science & Applied Physics (Imaging Photonics) | Intercept | 0.07061706 | 0.001299543 | 54.33990 | <0.0001 |
| | Female | −0.02074410 | 0.001517186 | −13.67274 | <0.0001 |
| Theoretical & Computational Sciences | Intercept | 0.09985401 | 0.001400810 | 71.28307 | <0.0001 |
| | Female | −0.02458395 | 0.001593115 | −15.43137 | <0.0001 |